\documentclass[conference]{IEEEtran}
\IEEEoverridecommandlockouts
\usepackage{cite}
\usepackage{amsmath,amssymb,amsfonts}
\usepackage{algorithmic}
\usepackage{graphicx}
\usepackage{textcomp}
\usepackage{xcolor}
\usepackage{fancyhdr}
\usepackage{eso-pic}

\usepackage[T1]{fontenc}
\usepackage[utf8]{inputenc}

\def\BibTeX{{\rm B\kern-.05em{\sc i\kern-.025em b}\kern-.08em
    T\kern-.1667em\lower.7ex\hbox{E}\kern-.125emX}}
\begin{document}

\fancyfoot{}
\fancyfoot[C]{\fontsize{8}{12} \selectfont © 2023 IEEE. Personal use of this material is permitted. Permission from IEEE must be obtained for all other uses, in any current or future media, including reprinting/republishing this material for advertising or promotional purposes, creating new collective works, for resale or redistribution to servers or lists, or reuse of any copyrighted component of this work in other works.}

\title{Explanation through Reward Model Reconciliation using POMDP Tree Search \\
\thanks{This material is based upon work supported by the Johns Hopkins University Applied Physics Laboratory and the Office of Naval Research.}
}

\makeatletter
\newcommand{\linebreakand}{%
  \end{@IEEEauthorhalign}
  \hfill\mbox{}\par
  \mbox{}\hfill\begin{@IEEEauthorhalign}
}
\makeatother

\author{
\IEEEauthorblockN{Benjamin D. Kraske\IEEEauthorrefmark{1}, Anshu Saksena\IEEEauthorrefmark{2}, Anna L. Buczak\IEEEauthorrefmark{2}, Zachary N. Sunberg\IEEEauthorrefmark{1}}
\IEEEauthorblockA{\textit{\IEEEauthorrefmark{1}Department of Aerospace Engineering Sciences, University of Colorado Boulder, Boulder, CO, USA}\\
\textit{\IEEEauthorrefmark{2}Applied Physics Laboratory, Johns Hopkins University, Laurel, MD, USA}\\
{\IEEEauthorrefmark{1}Benjamin.Kraske@colorado.edu, \IEEEauthorrefmark{2}Anshu.Saksena@jhuapl.edu, \IEEEauthorrefmark{2}Anna.Buczak@jhuapl.edu, \IEEEauthorrefmark{1}Zachary.Sunberg@colorado.edu}}
}

\maketitle \thispagestyle{fancy}

\AddToShipoutPictureBG*{%
  \AtPageUpperLeft{%
    \hspace{16.5cm}%
    \raisebox{-1.5cm}{%
      \makebox[0pt][r]{To Appear in the IEEE Int'l. Conference on Assured Autonomy (ICAA), 2023.}}}}
\begin{abstract}
    As artificial intelligence (AI) algorithms are increasingly used in mission-critical applications, promoting user-trust of these systems will be essential to their success. Ensuring users understand the models over which algorithms reason promotes user trust. This work seeks to reconcile differences between the reward model that an algorithm uses for online partially observable Markov decision (POMDP) planning and the implicit reward model assumed by a human user. Action discrepancies, differences in decisions made by an algorithm and user, are leveraged to estimate a user's objectives as expressed in  weightings of a reward function.
\end{abstract}

\begin{IEEEkeywords}
Explainabile Artificial Intelligence (XAI), Partially Observable Markov Decision Processes (POMDP), POMDP Planning
\end{IEEEkeywords}

\section{Introduction}
Artificial intelligence in the form of sequential decision making algorithms is increasingly used to address real world problems. These algorithms have many benefits, including the ability to reason over and account for future outcomes much more effectively than humans in many cases. However, these systems are only effective in real world applications if the decisions they recommend are trusted. In cases where algorithms and users have differing objectives, decisions and outcomes are likely to differ, which can lead to confusion, decreased trust in the system, and disuse of the system.


Consider resource allocation problems, where a limited number of resources must be allocated efficiently to number of needs. Examples of such problems are a space domain awareness sensor tasking problem, where a limited number of sensors are available to monitor multiple targets, or a repair dispatch problem, where limited repairpeople are available to make repairs at multiple locations. In such cases, we assume a user approves an allocation decision made by the algorithm. Here it is essential that the user understand why the algorithm arrives at its decision and that any misconceptions around the model or planning process be addressed before decision approval, ideally in real time.



This work seeks to reconcile differences between user and algorithm partially observable Markov decision (POMDP) models, focusing specifically on the objectives encoded in the reward model. We leverage action discrepancies between the algorithm and the user to estimate the user's reward model and provide explanations tailored to the user's misunderstanding of the model.


Section \ref{sec:bck} provides background on explainable artificial intelligence (XAI) and model reconciliation. Section \ref{sec:form} outlines the explanation problem. Section \ref{sec:soln} outlines our proposed solution. Section \ref{sec:result} provides an illustrative example, and Section \ref{sec:disc} contains discussion.

\section{Background \& Related Work} \label{sec:bck}

\subsection{POMDPs}
This work assumes familiarity with POMDPs \cite{kaelbling1998planning}, as defined by a tuple consisting of the state space $\mathcal{S}$, action space $\mathcal{A}$, transition model $\mathcal{T}$, observation space $\mathcal{O}$, observation model $\mathcal{Z}$, reward model $\mathcal{R}$, and a discount $\gamma$. 

\subsection{Explainable AI}

Explainable Artificial Intelligence (XAI) seeks to increase the transparency of algorithms by promoting user understanding of algorithm models and results. Although much of XAI focuses on machine learning and black-box methods there is an increasing focus on explainable planning \cite{chakraborti_emerging_2020}. 

The area of XAI most relevant to this work is introduced by Chakraborti et al. \cite{chakraborti_plan_2017}, who describe the model reconciliation problem, where explanations seek to align the model being reasoned over by a user with that of the planning algorithm. Sreedharan et al. \cite{ijcai2019p83} applied model reconciliation (for transition functions, reward functions, and discounts) to MDPs using a learning-based approach to determine which model parameters to explain. Tabrez et al. \cite{tabrez_explanation-based_2019} provide explanations of reward models for MDPs using an augmented POMDP model. Wang et al. \cite{wang_trust_2016} provide explanations of POMDP planning, but discussion of reward primarily involves presenting outcomes and values. Yadav et al. \cite{yadav_pomdps_2016} formulate a influence maximization problem as a POMDP and conduct user studies to determine users' reasoning over the networks which define the problem, building towards explanations of their solution.

Israelsen and Ahmed \cite{israelsen_davei_2019} provide a review of algorithmic assurances in human-autonomy trust. In particular, value-alignment or AI alignment, wherein user and algorithm goals are aligned, is discussed. Bobu et al. \cite{bobu_aligning_2023} review approaches to aligning human and robot representations of tasks, including reward learning. Yuan et al. \cite{yuan_situ_2022} develop a bi-directional approach to human-robot value alignment using a collaborative game formulation. Our work seeks to understand user goals so that relevant explanations of the current algorithm goals can be provided. 

\subsection{POMDP Inverse Reinforcement Learning}

Inverse reinforcement learning (IRL) seeks to learn objectives, as expressed in a reward function in the context of Markov decision processes, from expert trajectories. Numerous works have addressed POMDP inverse reinforcement learning \cite{choi_inverse_2011, chinaei_inverse_2012, chinaei_dialogue_2014, djeumou_task-guided_2022 }. Atrash and Pineau \cite{atrash_bayesian_2009} propose a framework for reinforcement learning using queries of actions from an optimal oracle to inform an estimate of the true reward distribution. Our work differs from these in that we coarsely estimate the reward function based on a single action and belief. Although in principle these IRL methods could be applied to our problem, we do not yet compare against them.
\section{Problem Statement} \label{sec:form}

Determining the minimal number of explanations to present to the user such that their model is updated while minimizing the exchange of information is a key challenge of model reconciliation \cite{ijcai2019p83}. Our work seeks to address model differences in the objectives as expressed in a POMDP reward function.

In many cases, the reward function $R$ of a POMDP can naturally be represented by a feature vector $\boldsymbol{\beta}(s,a)$ and a weighting $\phi$, such that $R(s,a) = \phi^T \boldsymbol{\beta}(s,a)$. We assume the user has an understanding of the reward features $\boldsymbol{\beta}(s,a)$, but differs in their valuing (or weighting) of each reward feature, $\phi$. Our work seeks to estimate user weightings and then use these weightings to provide concise explanations to the user which effectively update the user model. More formally, given a POMDP model $m$, a belief $b_\tau$, a planning algorithm action $a_{\phi_a,\tau}$, and a user-proposed alternative action $a_{\phi_h,\tau}$, we aim to find an estimate $\hat{\phi}_h$ of the user's weighting $\phi_h$.

\section{Solution} \label{sec:soln}
We use a proposed user action at a given timestep as a basis for estimating the user's reward weightings, $\hat{\phi}_h$, with a simple IRL-like scheme. Using this action, we estimate the user reward weighting through finding $\phi$ such that the user's proposed action has a higher estimated value than the algorithm action (fig. \ref{fig:rec}).
\begin{figure}[h]
    \centering
    \includegraphics[width = 0.4\textwidth]{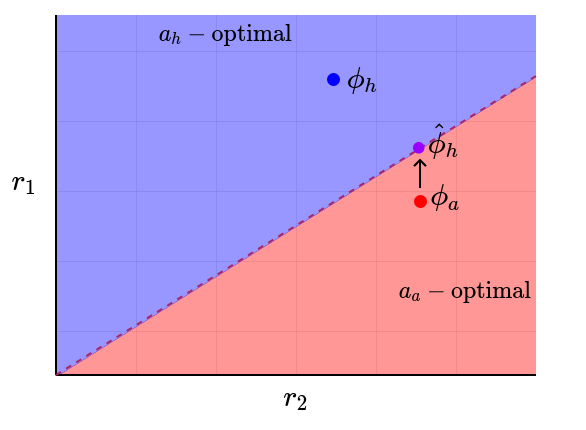}
    \caption{Estimating $\phi_h$ using action discrepancies}
    \label{fig:rec}
\end{figure}
\subsection{Constrained Optimization}
 We develop an explanation based on feedback from a single timestep in what we refer to as a one-shot approach. This approach enables explanations of discrepancies with a single user-proposed action rather than multiple timesteps or trajectories worth of user-proposed actions. The objective of the optimization problem is to find a reward weighting, \( \hat{\phi}_h\), for which the human action, $a_h$, is at least as good as the action chosen by the algorithm, $a_a$ while remaining as close as possible to original reward weighting $\phi_a$. This is mathematically formulated in the optimization problem below:
\begin{equation}\label{eqn:opt2}
\begin{aligned}
     &\underset{\hat{\phi}_{h}}{\text{minimize}}  \quad  \|\hat{\phi}_{h}-\phi_{a}\|_1 \\
     &\text{subject to }  Q^{\hat{\phi}_{h}}(b_\tau,a_{\phi_h,\tau}) \geq Q^{\hat{\phi}_{h}}(b_\tau,a^*_{\phi_a,\tau})\\
     & \qquad \qquad \qquad \qquad \quad \ \hat{\phi}_h \geq 0
\end{aligned}
\end{equation}
where $b_\tau$ is the belief at a given timestep $\tau$, $Q^{\hat{\phi}_{h}}(b,a)$ is belief-action value (evaluated on human reward weighting), $a_{{\phi}_h,\tau}$ is the user-proposed action at timestep $\tau$, and $a_{\phi_a,\tau}^*$ is the optimal action under the algorithm reward weighting at timestep $\tau$.



\subsection{Optimization}

The constrained optimization problem (\ref{eqn:opt2}) is reduced to an unconstrained optimization problem by penalizing constraint violations, yielding an objective function $U$:
\begin{equation}\label{eq:one_opt}
\begin{aligned}
    &U =\|\hat{\phi}_h-\phi_{a}\|_1+ wL_Q\\
    &\text{where } L_Q = -\max(|Q^{\hat{\phi}_h}(b_\tau,a^*_{\phi_a,\tau})- Q^{\hat{\phi}_h}(b_\tau,a_{\phi_h,\tau})|,0)
\end{aligned}
\end{equation}
with $w$ being a weighting variable in the relaxation which is ideally set such that the original constraint is not violated.

This relaxed optimization problem is solved using the Cross-Entropy method  \cite{rubinstein2004cross}, with the outputs restricted such that each element of $\phi$ is greater than 0. Cross-entropy provides a straightforward, gradient-free method for optimizing over the above loss functions, which are dependent on the DESPOT planner estimates of value which are calculated in real-time.

\subsection{POMDP Solutions}

In order to evaluate the loss function defined in (\ref{eq:one_opt}), estimates of the Q-values for the actions under $\phi$ are needed. An online POMDP solver, DESPOT \cite{ye_despot_2017}, is used to obtain Q-value estimates. This has the advantage of allowing our approach to scale to POMDPs with large state spaces that can only be solved in an online fashion.

\section{Illustrative Example} \label{sec:result}
\subsection{POMDP Formulation}
We formulate the following resource allocation problem, which provides an effective example on which to test explanations of decisions, especially those counterintuitive to would-be users. The HVAC (heating, ventilation, and air conditioning) repair dispatch problem involves deploying a repairpeople to one of several locations which may be ``ok'' or may be experiencing an HVAC fault (mechanical, electrical, or coolant fault). Locations are not always available, as customers may not always be at home. There is a cost for deploying repairpeople and if faults are not repaired within some time frame, a penalty is received. The problem is formulated as follows, with \( V=5\) being the horizon at which availability information is available, \(N=3\) the number of locations, \(R=2\) the number of repairpeople, and \(\textrm{T}=16\) the time horizon of the problem.

\noindent{\textbf{State Space}}

The state space $\mathcal{S}$ (\ref{eq:state}) is the product of the location statuses (including the time that the location status last changed), location availabilities, and timestep.
\begin{equation}\label{eq:state}
\begin{aligned}
    &\mathcal{S} = \mathrm{Status} \times \mathrm{Availability} \times [1,\mathrm{T}]\\
    &\mathrm{Status} = \{\{\mathsf{s}_i,\mathsf{t}_i\}_{i=1}^N \mid \mathsf{s}_i\in
     \{\mathrm{Ok, Mech., Elec., Cool.}\}, \\
    &\mathsf{t}_i\in[1,\mathrm{T}]\}\\
    &\mathrm{Availability} = \{\{\mathsf{al}_{vi}\}_{i=1}^N\mid\mathsf{al}_{vi}\in\{\mathrm{Avail, NA}\}, \\
    & v\in[1,V]\}
\end{aligned}
\end{equation}

\noindent{\textbf{Action Space}}

The action space $\mathcal{A}$ (\ref{eq:action}) is a tuple of length $R$ where each entry corresponds to the location where a repairperson is sent (location 0 indicates the repairperson is not sent out).
\begin{equation}\label{eq:action}
    \mathcal{A} = \{\{\mathsf{a}_{r}\}_{r=1}^R\mid\mathsf{a}_{r}\in[0,n]\}
\end{equation}

\noindent{\textbf{State Transition Distribution}}

The state transition model consists of three components corresponding to $\mathrm{Status}$, $\mathrm{Availability}$, and the overall timestep of the problem. Locations in one of the fault states ($\mathrm{Mech., Cool., Elec.}$) remain in that state unless a repairperson is sent to a location when it is available. Repairperson 1 specializes in coolant repairs. If they are sent to a location, mechanical faults are resolved with probability $0.8$, electrical faults are resolved with probability $0.9$, and coolant faults are resolved with probability $1.0$. If repairperson 2 is sent a location, any fault is resolved with probability $0.9$. A location's status remains $\mathrm{Ok}$ with probability $0.7$ and transitions to each of the fault states with probability $0.1$. If a location's status changes, the timestep portion of the status is reset to the current timestep.

\noindent{\textbf{Observation Space}}

The observation space $\mathcal{O}$ is defined in (\ref{eq:obs_space}).
\begin{equation}\label{eq:obs_space}
\begin{aligned}
    &\mathcal{O} = \mathrm{Status} \times \mathrm{Availability} \times [1,\mathrm{T}]\\
    & \mathrm{Status} = \{\{\mathsf{s}_i\}_{i=1}^N\mid_i\in\{\mathrm{Ok, Mech., Elec., Cool.}\}\}\\
    &\mathrm{Availability} = \{\{\mathsf{al}_{vi}\}_{i=1}^N\mid\mathsf{al}_{vi}\in\{\mathrm{Avail, NA}\},\\
    &v\in[1,V]\}
\end{aligned}
\end{equation}

\noindent{\textbf{Observation Distribution}}

The $\mathrm{Status}$ portion of the observation model provides noisy observations on the status. If a location is $\mathrm{Ok}$, an $\mathrm{Ok}$ observation is provided with probability $0.7$ and incorrect fault observations are provided with probability $0.1$ each. If a location is in a fault state, an accurate observation of that fault is provided with probability $0.5$, an incorrect $\mathrm{Ok}$ observation is provided with probability $0.1$, and incorrect observations of the other two faults are provided with probability $0.2$ each. The timestep at which the fault started is not directly observable.
The $\mathrm{Availability}$ of locations and the problem timestep are fully observable. 



\noindent{\textbf{Rewards}}

The reward function is defined using weighting terms $\phi$:
\begin{equation}\label{eq:rew}
\begin{aligned}
    &R(s,a) = \phi^T \boldsymbol{\beta}(s,a)\\
    &\text{where } \boldsymbol{\beta}(s,a)= \\
    &[ \bar{R}_{L1}(s,a), ... ,\bar{R}_{Ln}(s,a), \bar{R}_{W1}(s,a), ..., \bar{R}_{Wr}(s,a)]\\
    &\bar{R}_{Ln}(s,a) = 
    \begin{cases} 
        r_{l_{n}} \text{if Loc. $n$ status is fault for $x_{l_{n}}$}\\ \text{ \quad \; timesteps}\\
        0 \text{ otherwise}
    \end{cases}\\
    &\bar{R}_{Wr}(s,a) = 
    \begin{cases} 
        r_{w_{r}} \text{ if send repairperson $r$}\\
        0 \text{ otherwise}
    \end{cases}
\end{aligned}
\end{equation}

where $x_{l_{n}}$ and $r_{l_{n}}$ (a negative number) are set for each of the $N$ locations and $r_{w_r}$ (a negative number) is set for each of the $R$ repairpeople. By default, $\phi$ terms are $1$, but these terms may be changed in order to vary the relative weighting of penalties for locations and wages for repairpeople. The reward parameters for the example problem discussed are: $r_{l_1} = -250$, $r_{l_{2,3}} = -125$, $x_{l_{1,2,3}} = 3$, $r_{w_1} = -5$, and $r_{w_2} = -4$.



\subsection{Optimization Example}
\begin{figure*}[t]
    \centering
    \includegraphics[width = 0.85\textwidth]{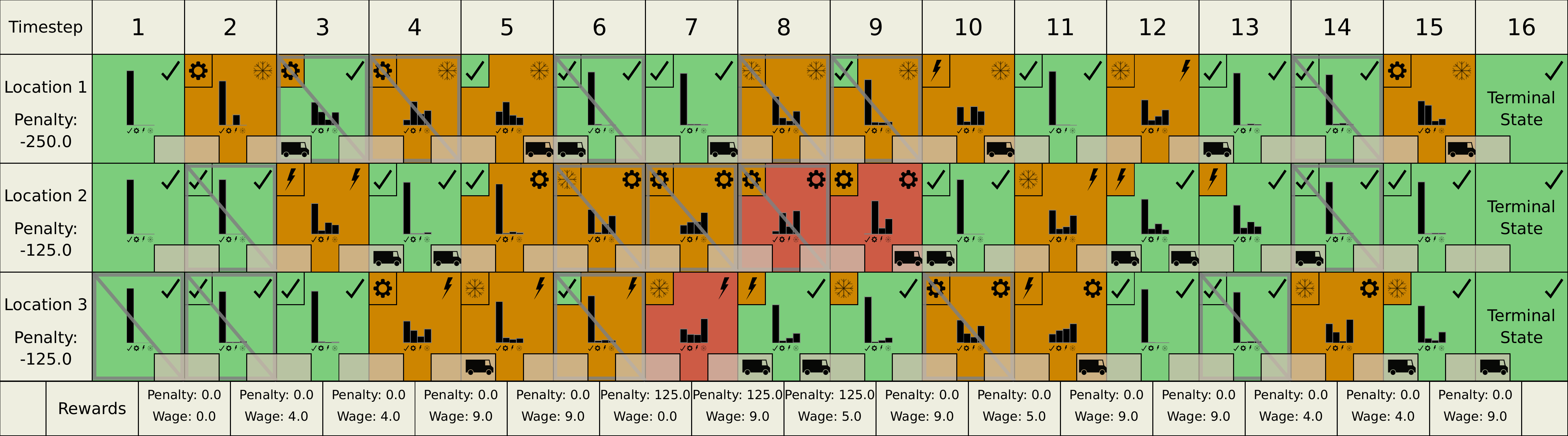}
    \caption{A simulation visualization. Note the penalties incurred at timesteps 7, 8, and 9 (shown in red). Observations are shown in the upper left of each cell, while the true state is in the upper right. The belief  over states is shown in the center and action taken in the lower right.}
    \label{fig:ex}
    \vspace{-5mm}
\end{figure*}

As a demonstration of the one-shot reward reconciliation, we consider a timestep from the simulation depicted in fig. \ref{fig:ex}, where a sub-optimal approximate algorithm is used to obtain a solution. The algorithm reward weighting, $\phi_a$, is defined $\phi_{a}=[1, 1, 1, 1, 1]$.


With this $\phi_a$, at timestep $5$, action $(1,1)$ is executed, sending both repairpeople to location 1. The user may wonder why the algorithm does not distribute the repairpeople more evenly across locations (which are all experiencing faults) and propose action $(2,1)$, sending repairperson 1 to location 2 and repairperson 2 to location 1. Given this user action, the algorithm action, and the belief at the timestep, the one-shot optimization routine is performed, estimating a point $\phi$ on the boundary in the space of weightings at which $Q^{\phi}(b,a_a) = Q^{\phi}(b,a_h)$ (see fig. \ref{fig:rec}). The following estimated human reward weighting $\hat{\phi}_h$ is returned and can be compared to $\phi_a$.
\begin{equation*}
    \hat{\phi}_{h}=[0.680, 1.007, 1.000, 0.999, 1.000]
\end{equation*}
where the first three elements correspond the penalty weightings for the three locations and the last two elements are the weightings for repairperson wages. 

Comparing these two weightings, the value of location 1 is reduced to roughly $70\%$ of the value in $\phi_a$. This difference in weightings now serves as the basis for a concise explanation of a portion of the problem objectives to the user. There are a number of potential means of presenting this information to the user. One approach to this explanation is to remind the user that, as currently formulated, location 1 has a higher penalty than the other locations, in a manner which conveys the estimated weighting:




\begin{quote}
    ``You seem to value the penalty at Location 1 at 70\% of what the algorithm does."
\end{quote}

Given this information, which the user may or may not have known or remembered, we can incorporate user feedback on the usefulness of this information and repeat this process as needed, with alternative actions  again proposed by the user. Although not in the scope of this work, this process could also be used to update the algorithm weighting if user-feedback on the objectives of the problem is desired.

\section{Conclusion} \label{sec:disc}
The above example provides a proof of concept for our proposed approach to model reconciliation for POMDPs. However, more thorough evaluation is needed. A challenge of this work, and XAI methods in general, is developing appropriate means of evaluation. Our continuing work focuses on means of evaluating and improving the work presented here.

Additionally, the reward weighting estimation problem, like many IRL problems, is underspecified. There may be many possible reward weightings which result in the same action for a given belief. Our proposed optimization problem, avoids this by minimizing the distance between $\phi_a$ and $\hat{\phi}_h$. However, given only one action and belief, it is difficult to precisely estimate the user's true reward weighting $\phi_h$. $\hat{\phi}_h$ does not approximate $\phi_h$ well in all cases, especially if $\phi_h$ is not close to $\phi_a$ (see fig. \ref{fig:rec}). Increasing the accuracy of $\hat{\phi}_h$, potentially through multiple rounds of explanation and feedback, is an area of future work.



\section*{Acknowledgment}
The authors thank Caroline Rogers and Michael Burke at the Johns Hopkins University Applied Physics Laboratory for their input and support.
\bibliographystyle{IEEEtran}
\bibliography{IEEEabrv,shorterbib}

\begin{thebibliography}{10}
\providecommand{\url}[1]{#1}
\csname url@samestyle\endcsname
\providecommand{\newblock}{\relax}
\providecommand{\bibinfo}[2]{#2}
\providecommand{\BIBentrySTDinterwordspacing}{\spaceskip=0pt\relax}
\providecommand{\BIBentryALTinterwordstretchfactor}{4}
\providecommand{\BIBentryALTinterwordspacing}{\spaceskip=\fontdimen2\font plus
\BIBentryALTinterwordstretchfactor\fontdimen3\font minus
  \fontdimen4\font\relax}
\providecommand{\BIBforeignlanguage}[2]{{%
\expandafter\ifx\csname l@#1\endcsname\relax
\typeout{** WARNING: IEEEtran.bst: No hyphenation pattern has been}%
\typeout{** loaded for the language `#1'. Using the pattern for}%
\typeout{** the default language instead.}%
\else
\language=\csname l@#1\endcsname
\fi
#2}}
\providecommand{\BIBdecl}{\relax}
\BIBdecl

\bibitem{kaelbling1998planning}
L.~P. Kaelbling, M.~L. Littman, and A.~R. Cassandra, ``Planning and acting in
  partially observable stochastic domains,'' \emph{Artificial intelligence},
  vol. 101, no. 1-2, pp. 99--134, 1998.

\bibitem{chakraborti_emerging_2020}
T.~Chakraborti, S.~Sreedharan, and S.~Kambhampati, ``The emerging landscape of
  explainable automated planning \& decision making,'' in \emph{Proc. 29th Int.
  Joint Conf. Artif. Intell., {IJCAI-20}}, C.~Bessiere, Ed., 7 2020, pp.
  4803--4811, survey track.

\bibitem{chakraborti_plan_2017}
T.~Chakraborti, S.~Sreedharan, Y.~Zhang, and S.~Kambhampati,
  ``\BIBforeignlanguage{en}{Plan {Explanations} as {Model} {Reconciliation}:
  {Moving} {Beyond} {Explanation} as {Soliloquy}},'' in
  \emph{\BIBforeignlanguage{en}{Proc. 26th Int. Joint Conf. Artif. Intell.}},
  Aug. 2017, pp. 156--163.

\bibitem{ijcai2019p83}
S.~Sreedharan, A.~O. Hernandez, A.~P. Mishra, and S.~Kambhampati, ``Model-free
  model reconciliation,'' in \emph{Proc. of the 28th Int. Joint Conf. on
  Artificial Intelligence, {IJCAI-19}}, 7 2019, pp. 587--594.

\bibitem{tabrez_explanation-based_2019}
A.~Tabrez, S.~Agrawal, and B.~Hayes, ``Explanation-{Based} {Reward} {Coaching}
  to {Improve} {Human} {Performance} via {Reinforcement} {Learning},'' in
  \emph{2019 14th {ACM}/{IEEE} {Int.} {Conf.} on {Human}-{Robot} {Interaction}
  ({HRI})}, Mar. 2019, pp. 249--257, iSSN: 2167-2148.

\bibitem{wang_trust_2016}
N.~Wang, D.~V. Pynadath, and S.~G. Hill, ``Trust calibration within a
  human-robot team: {Comparing} automatically generated explanations,'' in
  \emph{2016 11th {ACM}/{IEEE} {Int.} {Conf.} on {Human}-{Robot} {Interaction}
  ({HRI})}, Mar. 2016, pp. 109--116, iSSN: 2167-2148.

\bibitem{yadav_pomdps_2016}
A.~Yadav, H.~Chan, A.~Jiang, E.~Rice, E.~Kamar, B.~Grosz, and M.~Tambe,
  ``\BIBforeignlanguage{en}{{POMDPs} for {Assisting} {Homeless} {Shelters} –
  {Computational} and {Deployment} {Challenges}},'' in
  \emph{\BIBforeignlanguage{en}{Autonomous {Agents} and {Multiagent}
  {Systems}}}, ser. Lecture {Notes} in {Computer} {Science}, N.~Osman and
  C.~Sierra, Eds.\hskip 1em plus 0.5em minus 0.4em\relax Cham: Springer
  International Publishing, 2016, pp. 67--87.

\bibitem{israelsen_davei_2019}
B.~W. Israelsen and N.~R. Ahmed, ``\BIBforeignlanguage{en}{“{Dave}...{I} can
  assure you ...that it’s going to be all right ...” {A} {Definition},
  {Case} for, and {Survey} of {Algorithmic} {Assurances} in {Human}-{Autonomy}
  {Trust} {Relationships}},'' \emph{\BIBforeignlanguage{en}{ACM Computing
  Surveys}}, vol.~51, no.~6, pp. 1--37, Nov. 2019.

\bibitem{bobu_aligning_2023}
A.~Bobu, A.~Peng, P.~Agrawal, J.~Shah, and A.~D. Dragan, ``Aligning {Robot} and
  {Human} {Representations},'' Feb. 2023, arXiv:2302.01928 [cs].

\bibitem{yuan_situ_2022}
L.~Yuan, X.~Gao, Z.~Zheng, M.~Edmonds, Y.~N. Wu, F.~Rossano, H.~Lu, Y.~Zhu, and
  S.-C. Zhu, ``In situ bidirectional human-robot value alignment,''
  \emph{Science Robotics}, vol.~7, no.~68, p. eabm4183, Jul. 2022, publisher:
  American Association for the Advancement of Science.

\bibitem{choi_inverse_2011}
J.~Choi and K.-E. Kim, ``Inverse {Reinforcement} {Learning} in {Partially}
  {Observable} {Environments},'' \emph{Journal of Machine Learning Research},
  vol.~12, no.~21, pp. 691--730, 2011.

\bibitem{chinaei_inverse_2012}
H.~R. Chinaei and B.~Chaib-Draa, ``An {Inverse} {Reinforcement} {Learning}
  {Algorithm} for {Partially} {Observable} {Domains} with {Application} on
  {Healthcare} {Dialogue} {Management},'' in \emph{2012 11th {Int.} {Conf.}
  {Machine} {Learning} and {Applications}}, vol.~1, Dec. 2012, pp. 144--149.

\bibitem{chinaei_dialogue_2014}
H.~Chinaei and B.~Chaib-draa, ``\BIBforeignlanguage{en}{Dialogue {POMDP}
  components ({Part} {II}): learning the reward function},''
  \emph{\BIBforeignlanguage{en}{Int. Journal of Speech Technology}}, vol.~17,
  no.~4, pp. 325--340, Dec. 2014.

\bibitem{djeumou_task-guided_2022}
F.~Djeumou, M.~Cubuktepe, C.~Lennon, and U.~Topcu,
  ``\BIBforeignlanguage{en}{Task-{Guided} {Inverse} {Reinforcement} {Learning}
  under {Partial} {Information}},'' \emph{\BIBforeignlanguage{en}{Proc. Int.
  Conf. on Automated Planning and Scheduling}}, vol.~32, pp. 53--61, Jun. 2022.

\bibitem{atrash_bayesian_2009}
A.~Atrash and J.~Pineau, ``\BIBforeignlanguage{en}{A bayesian reinforcement
  learning approach for customizing human-robot interfaces},'' in
  \emph{\BIBforeignlanguage{en}{Proc. of the 14th Int. Conf. {Intelligent} User
  Interfaces}}.\hskip 1em plus 0.5em minus 0.4em\relax Sanibel Island Florida
  USA: ACM, Feb. 2009, pp. 355--360.

\bibitem{rubinstein2004cross}
R.~Y. Rubinstein and D.~P. Kroese, \emph{The cross-entropy method: a unified
  approach to combinatorial optimization, Monte-Carlo simulation, and machine
  learning}.\hskip 1em plus 0.5em minus 0.4em\relax Springer, 2004, vol. 133.

\bibitem{ye_despot_2017}
N.~Ye, A.~Somani, D.~Hsu, and W.~S. Lee, ``\BIBforeignlanguage{en}{{DESPOT}:
  {Online} {POMDP} {Planning} with {Regularization}},''
  \emph{\BIBforeignlanguage{en}{Journal of Artificial Intelligence Research}},
  vol.~58, pp. 231--266, Jan. 2017.

\end{thebibliography}


\end{document}